\documentclass[letterpaper, 10 pt, conference]{ieeeconf}  

\IEEEoverridecommandlockouts                              

\usepackage{graphicx}
\usepackage{amsmath,amssymb} 
\usepackage{color}
\usepackage{hyperref}
\usepackage{subfig}
\usepackage{tabularx}
\usepackage{amsfonts}
\usepackage{multirow}
\usepackage{flushend}

\newcommand{\T}[2]{\mathbf{T}^{#1}_{#2}}
\newcommand{\Pl}[1]{\boldsymbol{\pi}_{#1}}
\newcommand{\x}[1]{\mathbf{x}_{#1}}
\newcommand{\Q}[1]{\mathbf{Q}^{*}_{#1}}
\newcommand{\Qh}[1]{\mathbf{\hat{Q}}^{*}_{#1}}	  
\newcommand{\C}[1]{\mathbf{C}^{*}_{#1}}

\title{\LARGE \bf
Real-Time Monocular Object-Model Aware Sparse SLAM
}

\author{Mehdi Hosseinzadeh, Kejie Li, Yasir Latif, and Ian Reid
\thanks{All of the authors are with the Australian Center for Robotic Vision (ACRV) at the School of Computer Science, University of Adelaide {\tt\small \{firstname.lastname\}@adelaide.edu.au}.}
}

\begin{document}

\maketitle
\thispagestyle{empty}
\pagestyle{empty}

\begin{abstract}

Simultaneous Localization And Mapping (SLAM) is a fundamental problem in mobile robotics. While sparse point-based SLAM methods provide accurate camera localization, the generated maps lack semantic information. On the other hand, state of the art object detection methods provide rich information about entities present in the scene from a single image.
This work incorporates a real-time deep-learned object detector to the monocular SLAM framework for representing generic objects as quadrics that permit detections to be seamlessly integrated while allowing the real-time performance. Finer reconstruction of an object, learned by a CNN network, is also incorporated and provides a shape prior for the quadric leading further refinement.
To capture the dominant structure of the scene, additional planar landmarks are detected by a CNN-based plane detector and modeled as independent landmarks in the map.
Extensive experiments support our proposed inclusion of semantic objects and planar structures directly in the bundle-adjustment of SLAM - \textit{Semantic SLAM} - that enriches the reconstructed map semantically, while significantly improving the camera localization.
\end{abstract}


\section{INTRODUCTION}\label{sec:intro}

Simultaneous Localization And Mapping (SLAM) is one of the fundamental problems in
mobile robotics \cite{cadena2016past}
that aims to reconstruct a previously unseen environment while localizing a mobile robot with respect to it. 
The representation of the map is an important design choice as it directly affects its usability and precision.
A sparse and efficient representation for Visual SLAM is to consider the map as collection of points in 3D,
which carries information about geometry but not about the semantics of the scene.
Denser representations  \cite{dso,lsdslam,dtam,infinitam,kinectfusion}, remain equivalent to a collection of points in this regard.

Man-made environments contain many objects that can be used as landmarks in a SLAM map, encapsulating a higher level of abstraction than a set of points. Previous object-based SLAM efforts have mostly relied on a database of predefined objects -- which must be recognized and a precise 3D model fit to match the observation in the image to establish correspondence \cite{DBLP:conf/cvpr/Salas-MorenoNSKD13}. Other work \cite{Bao_CVPR2011_SSFM} has admitted more general objects (and constraints) but only in a slow, offline structure-from-motion context. In contrast, we are concerned with online (real-time) SLAM, but we seek to represent a wide variety of objects. Like \cite{Bao_CVPR2011_SSFM} we are not concerned with high-fidelity reconstruction of individual objects, but rather to represent the location, orientation and rough shape of objects, while incorporating fine point-cloud reconstructions on-demand. A suitable representation is therefore a quadric \cite{sfmquadric}, which captures a compact representation of rough extent and pose while allows elegant data-association. 
In addition to objects, much of the large-scale structure of a general scene (especially indoors) comprises dominant planar surfaces. 
Planes provide information complimentary to points by representing significant portions of the environment with few parameters, 
leading to a representation that can be constructed and updated online \cite{kaess-plane}.
In addition to constraining points that lie on them, planes
permit the introduction of useful affordance constraints between objects and their supporting surfaces that leads to better estimate of the camera pose. 

This work aims to construct a sparse semantic map representation consisting not only of points, but planes and objects as landmarks, all of which are used to localize the camera. 
We explicitly target real-time performance in a monocular setting 
which would be impossible with uncritical choices of representation and constraints. 
To that end, we use the representation for dual quadrics proposed in our previous work \cite{mehdi-arxiv} to represent and update general objects, 
\textbf{(1)} from front-end perspective such as:
\textbf{a)} reliance on the \textit{depth} channel for plane segmentation and parameter regression, 
\textbf{b)} pre-computation of Faster R-CNN \cite{fasterrcnn} based object detections to permit real-time performance, and
\textbf{c)} ad-hoc object and plane matching/tracking.  
\textbf{(2)} From the back-end perspective: 
\textbf{a)} conic observations are assumed to be axis-aligned thus limiting the robustness of the quadric reconstruction, 
\textbf{b)} all detected landmarks are maintained in a single global reference frame. 
This work in addition to addressing the mentioned limitations, proposes new factors amenable for real-time inclusion of plane and object detections while incorporating fine point-cloud reconstructions from a deep-learned CNN, wherever available, to the map and refine the quadric reconstruction according to this object model.

The main contributions of the paper as follows: 
(1) integration of two different CNN-based modules to segment planes and regress the parameters 
(2) integrating a real-time deep-learned object detector in a monocular SLAM framework to detect general objects as landmarks along a data-association strategy to track them, 
(3) proposing a new observation factor for objects to avoid axis-aligned conics, 
(4) representing landmarks relative to the camera where they are first observed instead of a global reference frame, and
(5) wherever available, integrating the reconstructed point-cloud model of the detected object from single image by a CNN to the map and imposing additional prior on the extent of the reconstructed quadric based on the reconstructed point-cloud. 

\section{RELATED WORK}\label{sec:relatedwork}
SLAM is well studied problem in mobile robotics and many different solutions have been proposed for solving it. 
The most recent of these is the graph-based approach that formulates SLAM as a nonlinear least squares problem 
\cite{grisetti2010tutorial}. 
SLAM with cameras has also seen advancement in theory and good implementations that have led to many real-time
systems from sparse (\cite{orbslam},\cite{dso}) to semi-dense (\cite{lsdslam}, \cite{svo}) to fully dense
(\cite{dtam}, \cite{kinectfusion}, \cite{infinitam}).

Recently, there has been a lot of interest in extending the capability of a point-based representation by either applying the same techniques to other geometric primitives or fusing points with lines or planes to get better accuracy. 
In that regard, \cite{kaess-plane} proposed a representation for modeling infinite planes and 
\cite{yang2016pop} use Convolutional Neural Network (CNN) to generate plane hypothesis from monocular images which are refined over time using both image planes and points. 
\cite{taguchi2013point} proposed a method to fuse points and planes from an RGB-D sensor. In the latter works, they try to fuse the information of planar entities to increase the accuracy of depth inference.

Quadrics based representation was first proposed in \cite{cross1998quadric} and later used in a structure from motion setup \cite{sfmquadric}. 
\cite{dualquadNiko2017arXiv} reconstructs quadrics based on bounding box detections, however it is not explicitly modeled to remain bounded ellipsoids. 
Addressing previous drawback, \cite{nicholson2019quadricslam} still relies on ground-truth data-association in a
non-real-time quadric-only framework.
\cite{sunderhauf2017meaningful} presented a semantic mapping system using object detection coupled with RGB-D SLAM, however object models do not inform localization. \cite{DBLP:conf/cvpr/Salas-MorenoNSKD13} presented an object based SLAM system that uses pre-scanned object models as landmarks for SLAM but can not be generalized to unseen objects.
\cite{mccormac2017semanticfusion} presented a system that fused multiple semantic predictions with a dense map reconstruction. SLAM is used as the backbone to establish multiple view correspondences for fusion of semantic labels but the semantic labels do not inform localization.

\section{Overview of the Landmark Representations and Factors}\label{sec:overview}
\begin{figure}[t]
\centering
\subfloat{\includegraphics[width=\columnwidth]{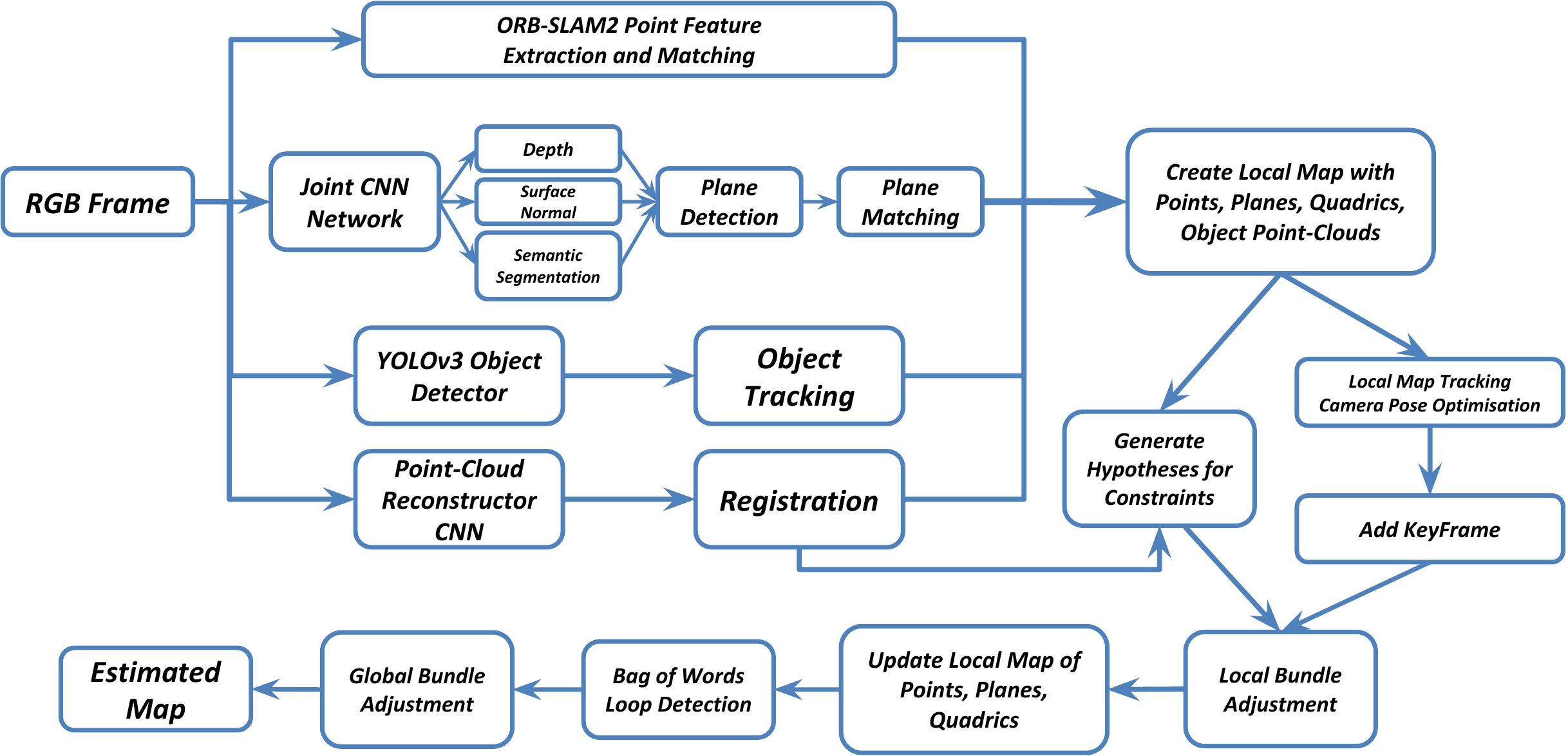}}\\
\caption{
The pipeline of our proposed SLAM system.
} 
\label{fig:system}
\end{figure}

For the sake of completeness, this section presents an overview of the representations and factors proposed originally in our previous work \cite{mehdi-arxiv}.
In the next sections, we propose new multi-edge observation and unary prior factors.
The SLAM problem can be represented as a bipartite factor~graph $\mathcal{G}(\mathcal{V},\mathcal{F},\mathcal{E})$ where $\mathcal{V}$ represents the set of \textit{vertices} (variables) that need to be estimated 
and $\mathcal{F}$ represents the set of \textit{factors} (constraints) that are connected to their associated variables by the set of edges $\mathcal{E}$. 
We propose our SLAM system in the context of factor~graphs.
The solution of this problem is the optimum configuration of vertices (MAP estimate), $\mathcal{V}^{*}$, that minimizes the overall error over the factors in the graph (log-likelihood of the joint probability distribution).
The pipeline of our SLAM system is illustrated in Fig~\ref{fig:system}.

\subsection{Quadric Representation}\label{subsec:overview_quadric}
A quadric surface in 3D space can be represented by a homogeneous quadratic form defined on the 3D projective space $ \mathbb{P}^{3} $ that satisfies $ \mathbf{x^{\top}Qx=0} $, where $ \mathbf{x} \in \mathbb{R}^{4} $ is the homogeneous 3D point and $ \mathbf{Q} \in \mathbb{R}^{4\times4} $ is the symmetric matrix representing the quadric surface. 
However, the relationship between a point-quadric $ \mathbf{Q} $ and its projection into an image plane (a conic) is not straightforward \cite{Hartley:2003:MVG:861369}. A widely accepted alternative is to make use of the dual space (\cite{cross1998quadric,sfmquadric,dualquadNiko2017arXiv}) which represents a dual quadric $ \Q{} $ by the envelope of planes $ \Pl{} $ tangent to it, viz: 
$ \Pl{}^{\top}\Q{}\Pl{}=0 $,
which simplifies the relationship between the quadric and its projection to a conic.
A dual quadric $ \Q{} $ can be decomposed as 
$ \Q{} = \T{}{Q} \Q{c} \T{\top}{Q} $
where $ \T{}{Q} \in \mathbf{SE}(3)$ transforms an axis-aligned (canonical) quadric at the origin, $ \Q{c} $, to a desired $ \mathbf{SE}(3) $ pose. 
Quadric landmarks need to remain bounded, i.e. ellipsoids, which requires $\Q{c}$ to have 3 positive and 1 negative eigenvalues. 
In \cite{mehdi-arxiv} we proposed a decomposition and incremental update rule for dual quadrics that guarantees this conditions and provides a good approximation for incremental update.
More specifically, the dual ellipsoid $ \Q{} $ is represented as a tuple $ \mathbf{(T,L)} $ where $ \mathbf{T \in SE}(3) $ and $ \mathbf{L} $ lives in $ \textbf{D}(3) $ the space of real diagonal $ 3 \times 3 $ matrices, i.e. an axis-aligned ellipsoid accompanied by a rigid transformation. 
The proposed approximate update rule for $ \Q{} = \mathbf{(T,L)} $ is: 
\begin{equation}
\resizebox{0.9\columnwidth}{!}{$
\mathbf{\Q{} \oplus \varDelta\Q{} = (T,L) \oplus (\varDelta T, \varDelta L) = (T \cdot \varDelta T, L + \varDelta L)}
$}
\end{equation}
where $ \mathbf{\oplus:\mathbb{E}\times\ \mathbb{E} \longmapsto \mathbb{E}} $ is the mapping for updating ellipsoids, $ \mathbf{\varDelta L} $ is the update for $ \mathbf{L} $ 
and $ \mathbf{\varDelta T} $  is the update for $ \mathbf{T} $ that 
are carried out in the corresponding lie-algebra of $ \mathfrak{d}(3) $ (isomorphic to $ \mathbb{R}^3 $) and $ \mathfrak{se}(3) $, respectively.

\subsection{Plane Representation}
Following \cite{kaess-plane}, a plane $ \Pl{} $ as a structural entity in the map is represented minimally by its normalized homogeneous coordinates $ \Pl{} = (a,b,c,d)^\top $ where $ \mathbf{n} = (a,b,c)^\top $ is the normal vector and $d$ is the signed distance to origin. 

\subsection{Constraints between Landmarks}\label{subsec:overview_constraints}

In addition to the classic point-camera constraint formed by the observation of a 3D point as 2D feature point in the camera, we model constraints between higher level landmarks and their observations in the camera. These constraints also carry semantic information about the structure of the scene, such as Manhattan assumption and affordances. 
We present a brief overview of these constraints here. In the next sections we present the newly introduced factors regarding plane and object observations and object shape priors, induced by the single-view point-cloud reconstructions.

\subsubsection{Point-Plane Constraint}
For a point $ \mathbf{x} $ to lie on its associated plane $ \Pl{} $ with the unit normal vector $\mathbf{n}$, we introduce the following factor between them:
\begin{equation}
f_{d}(\mathbf{x}, \Pl{})={{\parallel \mathbf{n}^{\top}(\mathbf{x}-\x{o}) \parallel}_{\sigma_d}^{2}}
\end{equation}
which measures the orthogonal distance of the point and the plane, 
for an arbitrary point  $ \x{o} $ on the plane.
$\|\mathbf{e}\|_{\boldsymbol{\Sigma}} $ notation is the Mahalanobis norm of $ \mathbf{e} $ and is defined as $ \mathbf{e}^\top \boldsymbol{\Sigma}^{-1}\mathbf{e}$ where $\boldsymbol{\Sigma}$ is the associated covariance matrix.

\subsubsection{Plane-Plane Constraint (Manhattan assumption)}
Manhattan world assumption where planes are mostly mutually parallel or perpendicular, is modeled as: 
\begin{align}
f_{\parallel}(\Pl{1}, \Pl{2}) & = {\parallel | \mathbf{n}_{1}^\top \mathbf{n}_{2} | - 1 \parallel}_{\sigma_{par}}^{2}   \qquad \text{\footnotesize{\it for parallel planes}} \\
f_{\perp}(\Pl{1}, \Pl{2}) & = {\parallel \mathbf{n}_{1}^\top \mathbf{n}_{2} \parallel}_{\sigma_{per}}^{2}       \qquad \text{\footnotesize{\it for perpendicular planes}}
\end{align}
where $ \Pl{1} $ and $ \Pl{2} $ have unit normal vectors $ \mathbf{n}_{1} $ and $ \mathbf{n}_{2} $.

\subsubsection{Supporting/Tangency Constraint}
In normal situations planar structure of the scene affords stable support for common objects, for instance floors and tables support indoor objects and roads support outdoor objects like cars. To impose a supporting affordance relationship between planar entities of the scene and common objects, we introduce a factor between dual quadric object $ \Q{} $ and plane $ \Pl{} $ that models the tangency relationship as: 
\begin{equation}
f_t(\Pl{}, \Q{})={\parallel \Pl{}^{\top}\Qh{}\Pl{} \parallel}_{\sigma_t}^{2}
\end{equation}
where $\Qh{}$ is the normalized dual quadric by its matrix Frobenius norm.
Please note that this tangency constraint is the direct consequence of choosing dual space for quadric representation, 
which is not straight-forward in point space.

%


\section{MONOCULAR PLANE DETECTION}\label{sec:mono_plane}

Man-made environments contain planar structures, such as table, floor, wall, road, etc. 
If modeled correctly,
they can provide information about large feature-deprived regions providing more map coverage. In addition, these landmarks act as a regularizer for other landmarks when constraints are introduced between them. 
The dominant approach for plane detection is to extract them from RGB-D input \cite{mehdi-arxiv} which provides reliable detection and estimation of plane parameters. 
In a monocular setting, planes need to be detected using a single RGB image and their parameters estimated, which is an ill-posed problem.
However, recent breakthroughs enable us to detect and estimate planes. 
Recently, PlaneNet~\cite{plane-net} presented a deeply learned network to predict plane parameters and corresponding segmentation masks. 
While planar segmentation masks are highly reliable, the regressed parameters are not accurate enough for small planar regions in indoor scenes (see Section \ref{sec:experiments}).
To address this shortcoming,
we use a network that predicts depth, surface normals, and semantic segmentations. Depth and surface normal contain complementary information about the orientation and distance of the planes, while semantic segmentation allows reasoning about identity of the region such as wall, floor, etc. 

\subsection{Planes from predicted depth, surface normals, and semantic segmentation}\label{subsec:plane_detection}
We utilize the state-of-the-art joint network~\cite{vlad-arxiv} to estimate depth, normals, and segmentation for each RGB frame in real-time. 
We exploit the redundancy in the three separate predictions 
to boost the robustness of the plane detection by generating plane hypotheses in two ways:
\textbf{1)} for each planar region in the semantic segmentation (regions such as floor, wall, etc.) we fit 3D planes using surface normals and depth for orientation and distance of the plane respectively, and
\textbf{2)} depth and surface normals predictions are utilized in the connected component segmentation of the reconstructed point-cloud in a parallel thread (\cite{trevor2013efficient, mehdi-arxiv}).
Plane detection 
$ \Pl{} = (a,b,c,d)^{\top} $ 
is considered to be valid if the cosine distance of normal vectors 
$ \mathbf{n} = (a,b,c)^{\top} $ 
and also the distance between the $ d $ value of the two planes from two estimations are within a certain threshold.
The corresponding plane segmentation is taken to be the intersection of the plane masks of the two hypotheses. 

Note that the association between 3D point landmarks and planes, useful for the factor described in \ref{subsec:overview_constraints}, is extracted from the resulting mask.
The 3D point is considered as an inlier if the corresponding 2D keypoint inside the mask also satisfies the certain geometric distance threshold.

\subsection{Plane Data Association}\label{subsec:plane_tracking}
Once initialized and added to the map, the landmark planes need to be associated with the detected planes in the incoming frames. 
Matching planes is more robust than feature point matching due to the inherent geometrical nature of planes \cite{mehdi-arxiv}.
To make data association more robust in cluttered scenes,
when available, we additionally use the detected keypoints that lie inside the segmented plane in the image to match the observations. 
A plane in the map and a plane in the current frame are deemed to be a match if 
the number of common keypoints is higher than a threshold $ th_H $ and the unit normal vector and distance of them are within certain threshold.
If the number of common keypoints is less than another threshold $ th_L $ (or zero for feature-deprived regions) meaning that there is no corresponding map plane for the detected plane, the observed plane is added to the map as a new landmark.  
The map can now contain two or more planar regions that might belong to the same infinite plane such as two tables with same height in the office. However, additional constraints on parallel planes are also introduced according to evidence (Section \ref{subsec:overview_constraints}).

\subsection{Multi-Edge Factor for Plane Observation}\label{subsec:plane_factor}
After successful data association, we can introduce the observation factor between the plane and the camera (keyframe). 
We use a relative key-frame formulation (instead of the global frame) for each plane landmark $ \Pl{r} $ which is expressed relative to the first key-frame ($ \T{w}{r} $) that observes it. 
For an observation $ \Pl{obs} $ from a camera pose $ \T{w}{c} $, the multi-edge factor (connected to more than two nodes) for measuring the plane observation is given by:
\begin{equation}
f_{\pi}(\Pl{r}, \T{w}{r}, \T{w}{c}) = {{\parallel d({\T{r}{c}}^{-\top}{\Pl{r}} , \Pl{obs}) \parallel}_{\boldsymbol{\Sigma}_\pi}^{2}}
\end{equation}
where $ {\T{r}{c}}^{-\top} \Pl{r} $ is the transformed plane from its reference frame to the camera coordinate frame and $ d $ is the geodesic distance of the $\mathbf{SO}(3)$ \cite{kaess-plane} and $\mathbf{T}_{c}^{w}$ is the pose of the camera which takes a point in the current camera frame ($\x{c}$) to a point in the world frame $\x{c} = {\mathbf{T}_{c}^{w}}{\x{w}}$. 

\section{Incorporating Object with Point-Cloud Reconstruction}\label{sec:objects}
As noted earlier, incorporating general objects in the map as quadrics leads to a compact representation of the rough 3D extent and pose (location and orientation) of the object while facilitating elegant data association. 
State-of-the-art object detector such as YOLOv3~\cite{yolov3} can provide object labels and bounding boxes in real-time for general objects.
The goal of introducing objects in SLAM is both to increase the accuracy of the localization and to yield a richer semantic map of the scene. 
While our SLAM proposes a sparse and coarse realization of the objects, wherever the fine model reconstruction of each object is available it can be seamlessly incorporated on top of the corresponding quadric and even refines the quadric reconstruction as discussed in \ref{subsec:obj_pointcloud}.

\subsection{Object Detection and Matching}\label{subsec:obj_match}

\begin{figure}[t]
\centering
\includegraphics[width=0.95\columnwidth]{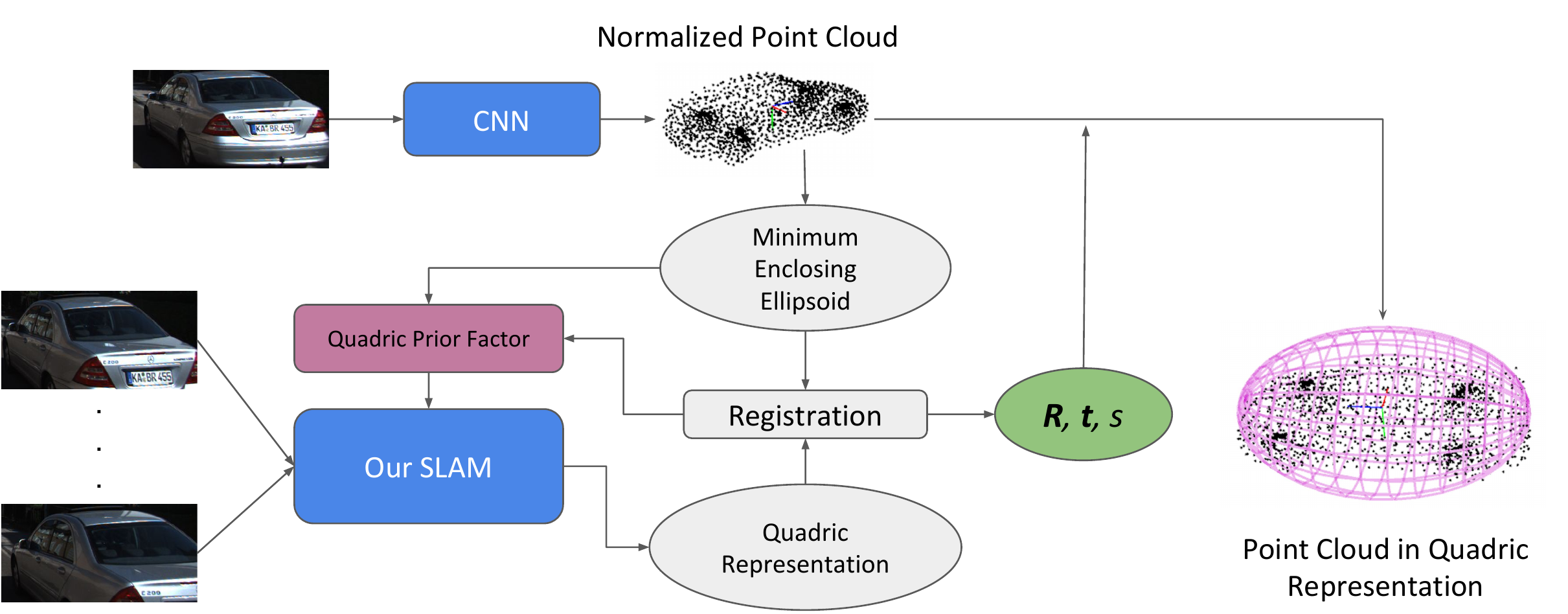}
\caption{Single-view point-cloud reconstruction imposes a shape prior constraint on a multi-view reconstructed quadric in our system (See Section \ref{subsec:obj_pointcloud})} 
\label{fig:pointcloud}
\end{figure} 

For real-time detection of objects, we use YOLOv3~\cite{yolov3} trained on COCO dataset~\cite{coco} that provides axis detections as aligned bounding boxes for common objects. 
For reliability we consider detections with 85\% or more confidence. 

\subsubsection*{Object Matching}
To rely solely on the geometry of the reconstructed quadrics (by comparing re-projection errors) to track the object detections against the map is not robust enough particularly for high-number of overlapping or partially-occluded detections.
Therefore to find optimum matches for all the detected objects in current frame, we solve the classic optimum assignment problem with Hungarin/Munkres~\cite{hungarian} algorithm. The challenge of using this classic algorithm is how to define the appropriate cost matrix.
We establish the cost matrix of this algorithm based on the idea of maximizing the number of common robustly matched keypoints (2D ORB features) inside the detected bounding boxes. Since we want to solve the minimization problem, the cost matrix is defined as: 
\begin{align}
    \mathbf{C} & = \left[c_{ij}\right]_{N \times M} \\
    c_{ij} & = K - p(b_i,q_j)
\end{align}
where $ p(b_i,q_j) $ gives the number of projected keypoints associated with candidate quadric $ q_j $ inside the bounding box $ b_i $, and $ K = \max_{\substack{i,j}} p(b_i,q_j) $ is the maximum number of all of these projected keypoints. $ N $ and $ M $ are the total number of bounding box detections in current frame and candidate quadrics of the map for matching, respectively. Candidate quadrics for matching are considered to be the quadrics of the map that are currently in front of the camera.

To reduce the number of mismatches furthermore, after solving the assignment problem with the proposed cost matrix, the solved assignment of $ b_i^* $ to $ q_j^* $ is considered successful if the number of common keypoints satisfies a certain high threshold $  p(b_i^*,q_j^*) \geq th_{high} $ and the new quadric will be initialized in the map if $  p(b_i^*,q_j^*) \leq th_{low} $. Assignments with  $ p(b_i^*,q_j^*) $ values between these thresholds will be ignored.

\subsection{Point-Cloud Reconstruction and Shape Priors}\label{subsec:obj_pointcloud}

In this section, we present a method of estimating fine geometric model of available objects established on top of quadrics to enrich their inherent coarse representation.
It is difficult to estimate the full 3D shape of objects from sparse views using purely classic geometric methods. To bypass this limitation,
we train a CNN adapted from Point Set Generation Net \cite{fan2017point} to predict (or hallucinate) the accurate 3D shape of objects as point clouds from single RGB images.

The CNN is trained on a CAD model repository ShapeNet \cite{chang2015shapenet}. We render 2D images of CAD models from random viewpoints and, to simulate the background in real images, we overlay random scene backgrounds from the SUN dataset \cite{sundataset} on the rendered images. 
We demonstrate the efficacy of this approach for outdoor scenes, particularly for general car objects in KITTI~\cite{kitti} benchmark in section \ref{subsec:kitti}.
Running alongside with the SLAM system, the CNN takes an amodal detected bounding box of an object as input and generates a point cloud to represent the 3D shape of the object. However, to ease the training of the CNN, the reconstructed point cloud is in a normalized scale and canonical pose. To incorporate the point cloud into the SLAM system, we need to estimate seven parameters to scale, rotate and translate this point cloud. First we compute the minimum enclosing ellipsoid of the normalized point cloud, and then estimate the parameters by aligning it to the object ellipsoid from SLAM.

\subsubsection*{Shape Prior on Quadrics}
After registering the reconstructed point-cloud and the quadric from SLAM, we impose a further constraint only on the shape (extent) of the quadric, Fig~\ref{fig:pointcloud}, feasible due to the decomposition of quadric representation. This prior affects the ratio of major axes of the quadric $ \Q{} $ by computing the intersection over union of the registered enclosing normalized cuboid of the point-cloud $ \mathcal{M} $ and enclosing normalized cuboid of the quadric:
\begin{equation}
\resizebox{0.9\columnwidth}{!}{$
f_{prior}(\Q{}) = {\| 1 - IoU_{cu}(cuboid(\Q{}), cuboid(\mathcal{M})) \|_{\sigma_p}^2}
$}
\end{equation}
where $ cuboid $ is a function that gives the normalized enclosing cuboid of an ellipsoid.

As an expedient approach, we currently pick a single high-quality detected bounding box as the input to the CNN, however, it is not complicated to extend to multiple bounding boxes by using a Recurrent Neural Net to fuse information from different bounding boxes, as done in 3D-R2N2 \cite{choy20163d}.

\subsection{Multi-Edge Factor for Non-Aligned Object Observation}\label{subsec:obj_factor}
We propose an observation factor for the quadric without enforcing that to be observed as an axis-aligned inscribed conic (ellipse). 
Unlike \cite{dualquadNiko2017arXiv} that uses the Mahalanobis distance of detected and projected bounding boxes, which is not robust and penalizes more for large errors and outliers, we use the error function based on Intersection-over-Union (IoU) of these bounding boxes that is also weighted according to the \textit{confidence score} $s$ of the object detector. 
This factor provides an inherent capped error, however it implicitly emphasizes on the significance of the good initialization of quadrics to have a successful optimization.
Similar to plane landmarks, we use the relative reference key-frame $ \T{w}{r} $ to represent the coordinates of the objects, we introduce the multi-edge factor, for object observation error, between dual quadric $ \Q{r} $ and camera pose $ \mathbf{\T{w}{c}} $ as:
\begin{equation}
f_Q(\Q{r}, \T{w}{r}, \T{w}{c})=
\parallel 1 - IoU_{bb}(B^{*} , B_{obs}) \parallel_{s^{-1}}^2
\end{equation}
where $ B_{obs} $ is the detected bounding box and $ B^{*} $ is the enclosing bounding box of the projected conic $ \C{} \sim \mathbf{P} \Q{r} \mathbf{P}^\top $ with the projection matrix $ \mathbf{ P = K} \begin{bmatrix} \mathbf{I}_{3\times3} & \mathbf{0}_{3\times3} \end{bmatrix} \T{r}{c} $ of the camera with calibration matrix $ \mathbf{K} $ ,\cite{Hartley:2003:MVG:861369}, and $ \T{r}{c} = {\T{w}{c}}{(\T{w}{r})}^{-1} $ is the relative pose of the camera from the reference key-frame of the quadric.

\section{EXPERIMENTS}\label{sec:experiments}

\begin{figure}[t]
\centering
\subfloat{\includegraphics[width=0.11\textwidth]{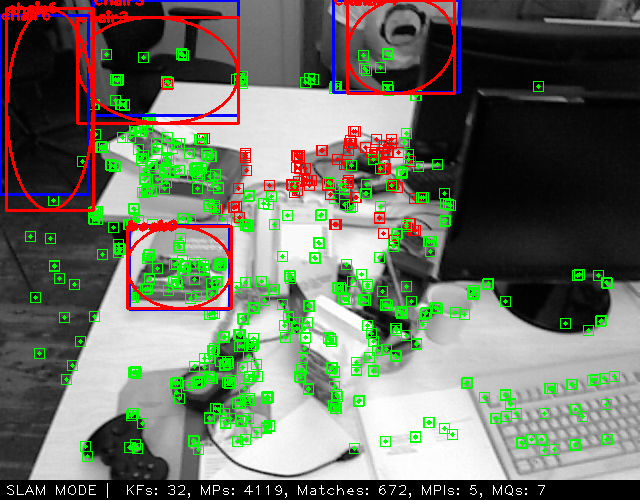}}~
\subfloat{\includegraphics[width=0.11\textwidth]{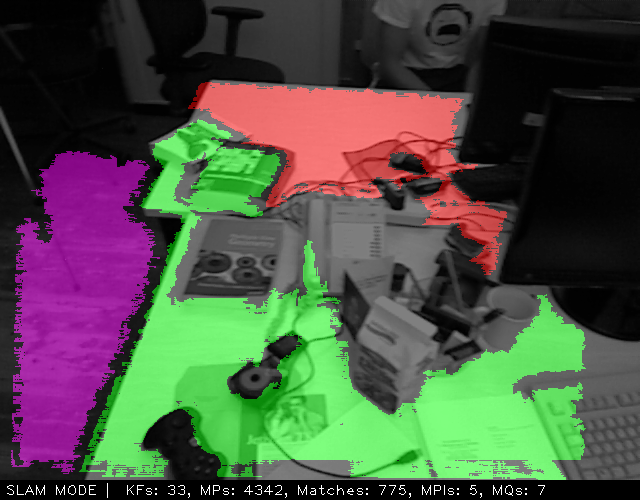}}~
\subfloat{\includegraphics[width=0.11\textwidth]{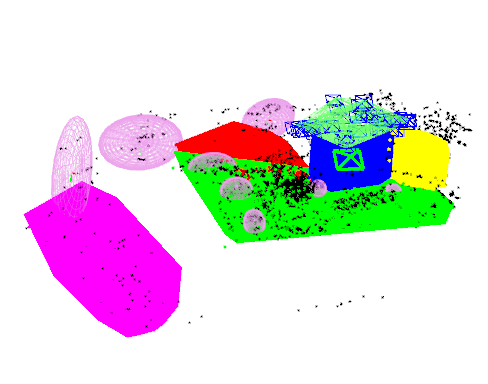}}~
\subfloat{\includegraphics[width=0.11\textwidth]{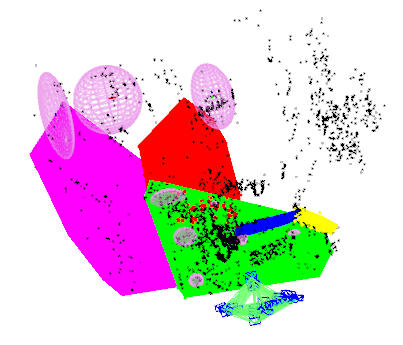}} \\
\vspace{-2mm}
\subfloat{\includegraphics[width=0.11\textwidth]{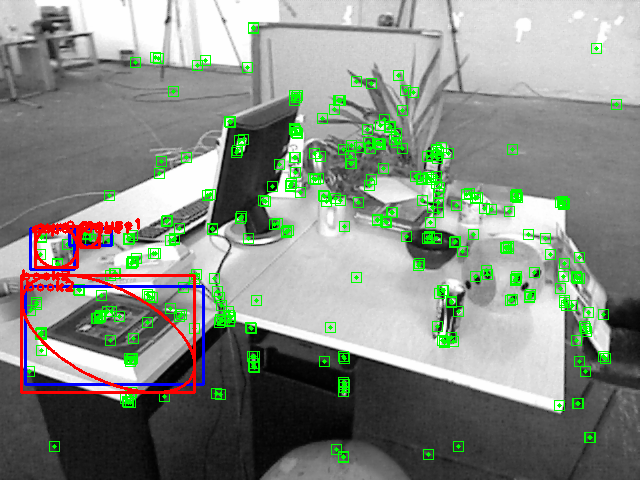}}~
\subfloat{\includegraphics[width=0.11\textwidth]{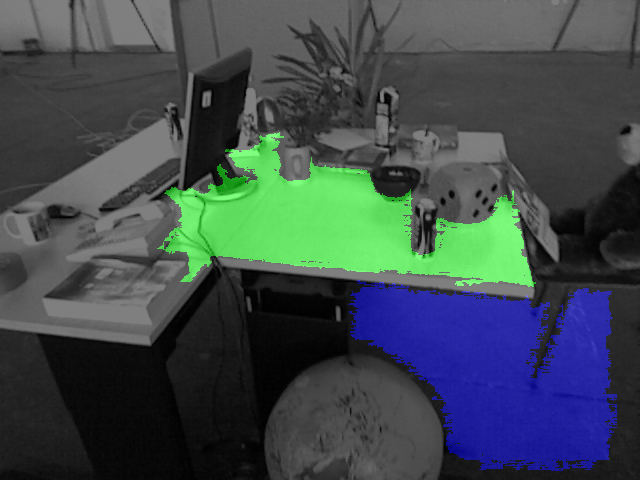}}~
\subfloat{\includegraphics[width=0.11\textwidth]{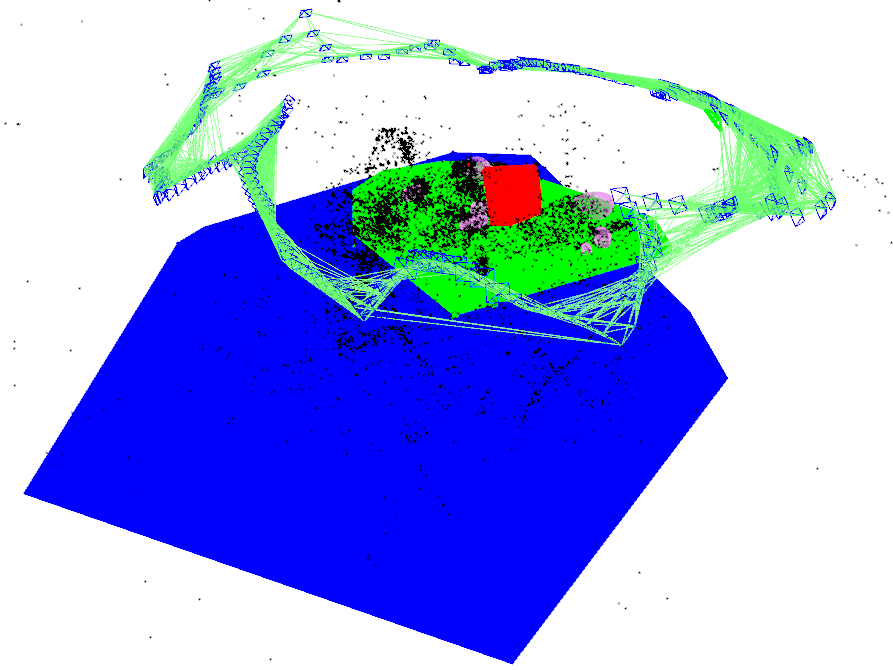}}~
\subfloat{\includegraphics[width=0.11\textwidth]{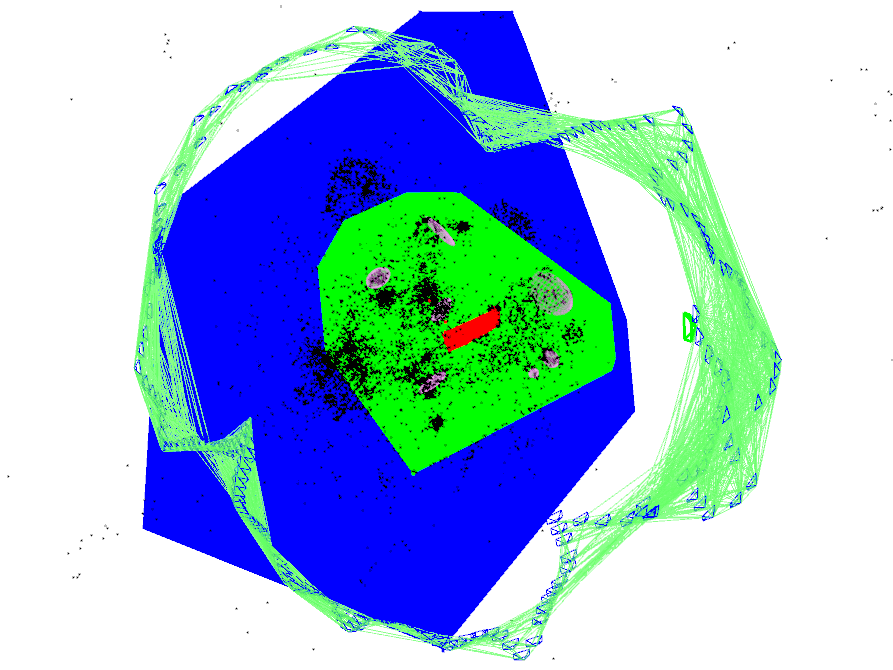}}\\
\vspace{-2mm}
\subfloat{\includegraphics[width=0.11\textwidth]{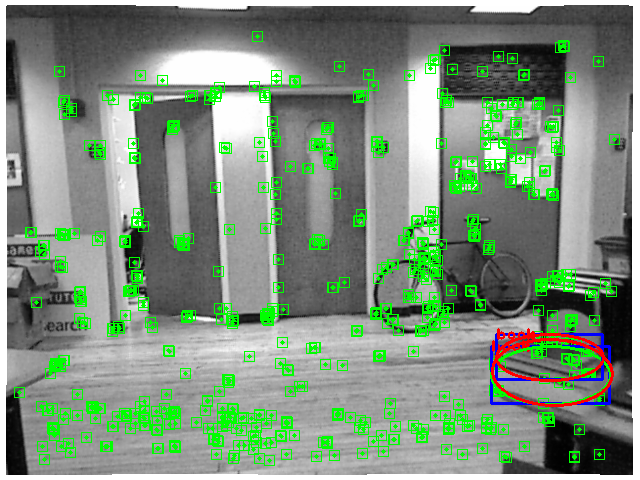}}~
\subfloat{\includegraphics[width=0.11\textwidth]{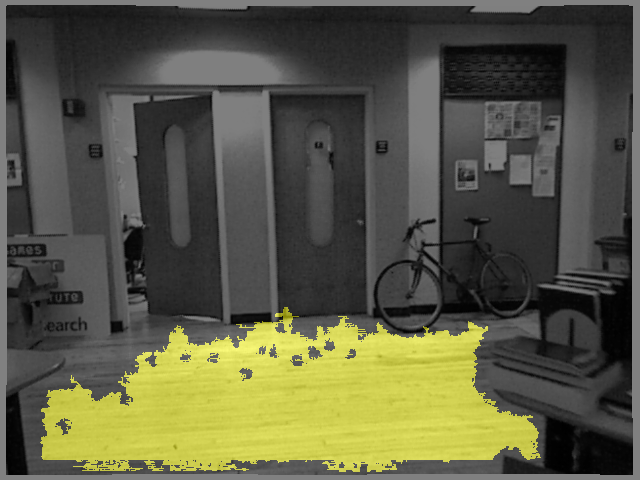}}~
\subfloat{\includegraphics[width=0.11\textwidth]{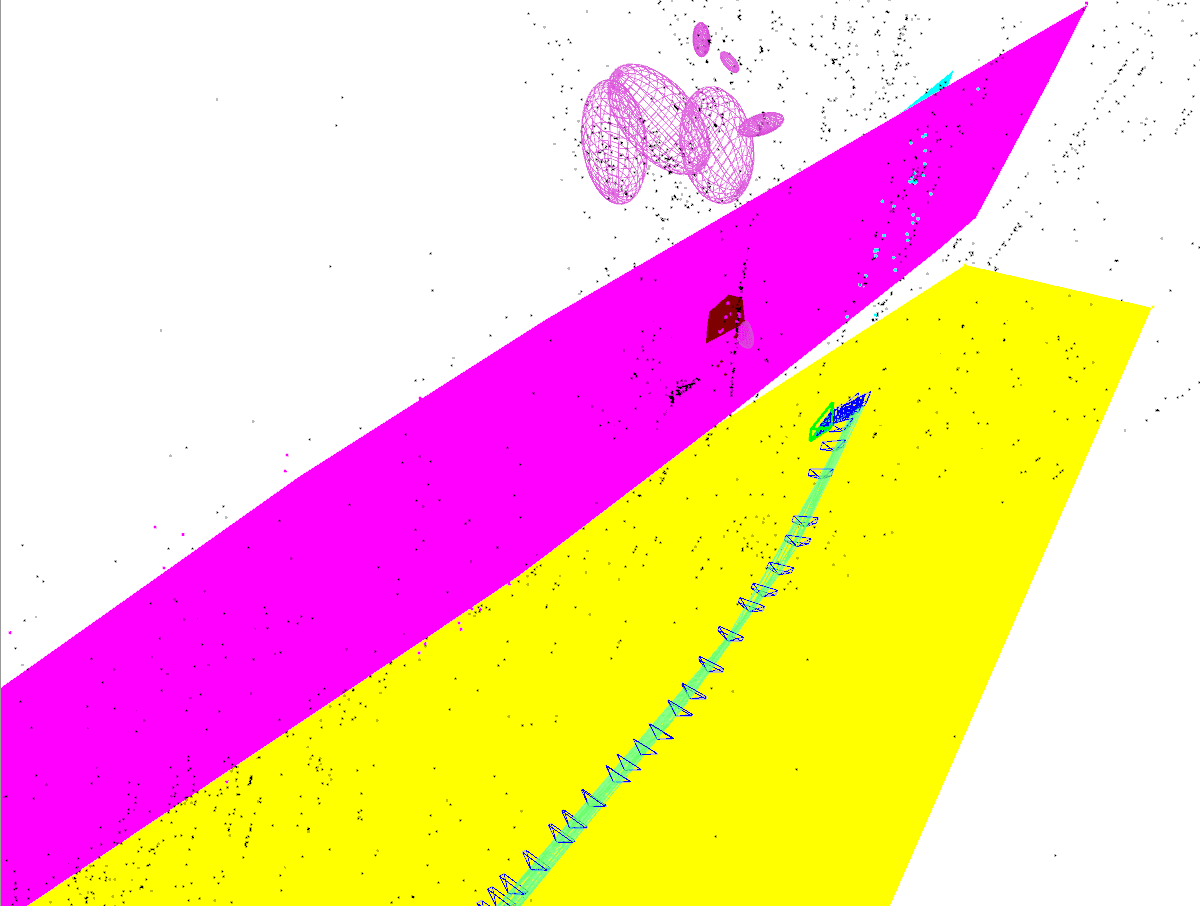}}~
\subfloat{\includegraphics[width=0.11\textwidth]{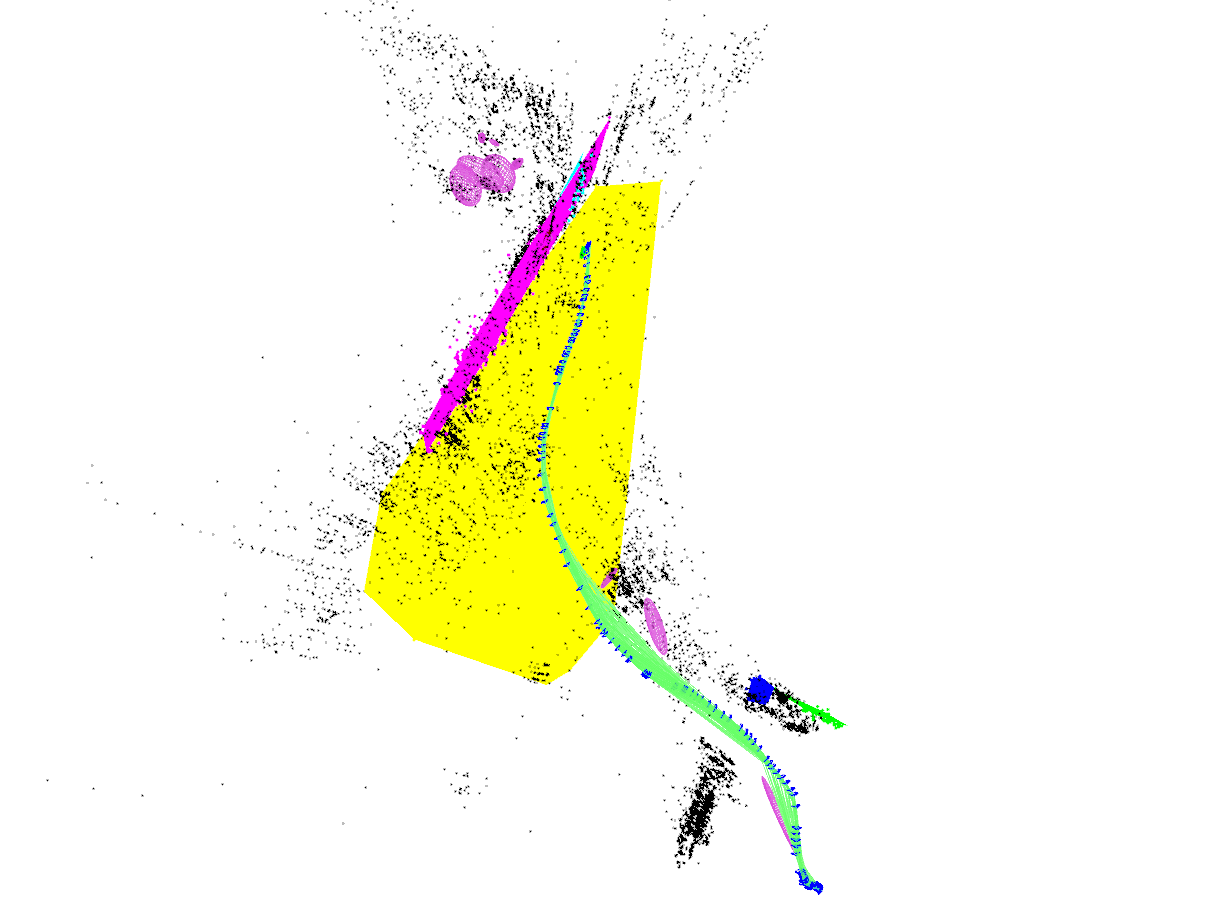}}\\
\vspace{-2mm}
\setcounter{subfigure}{0}
\subfloat[ORB Features and Detected Objects]{\includegraphics[width=0.11\textwidth]{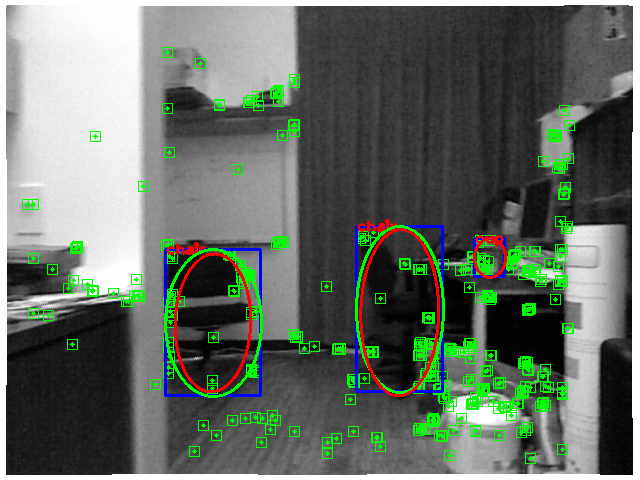}}~
\subfloat[Segmented Planes]{\includegraphics[width=0.11\textwidth]{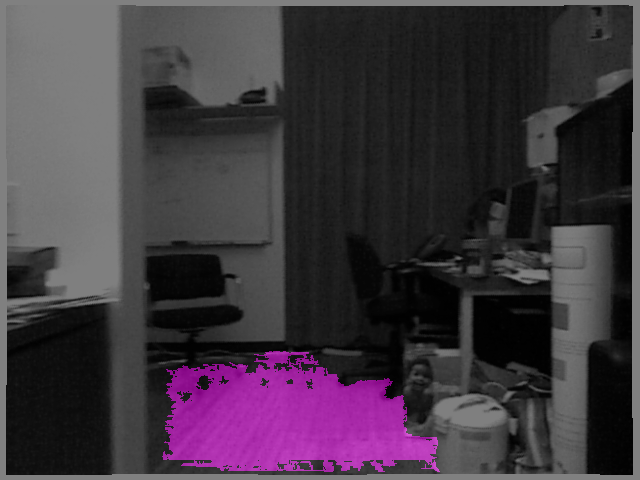}}~
\subfloat[Generated Map (Side)]{\includegraphics[width=0.11\textwidth]{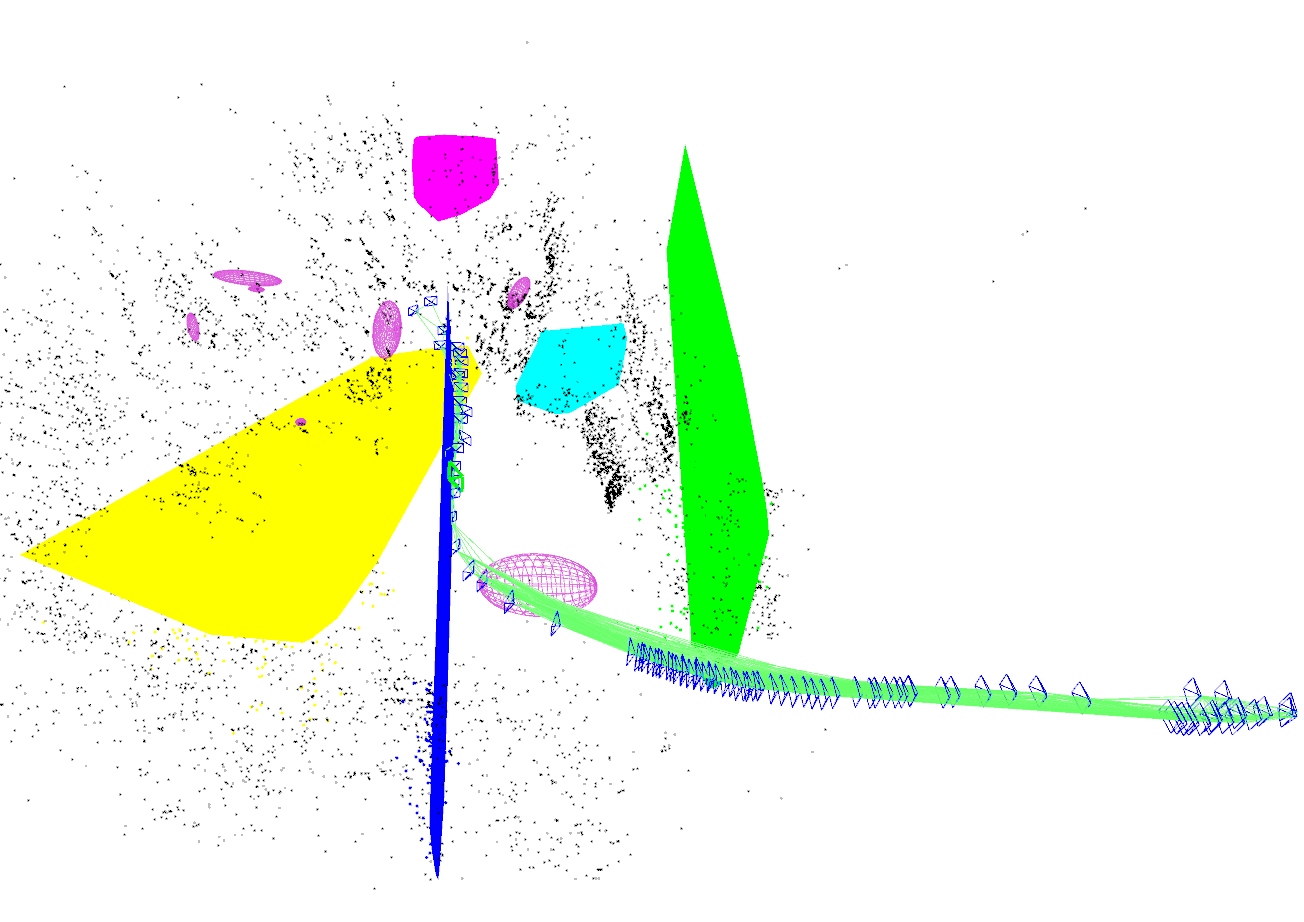}}~
\subfloat[Generated Map (Top)]{\includegraphics[width=0.11\textwidth]{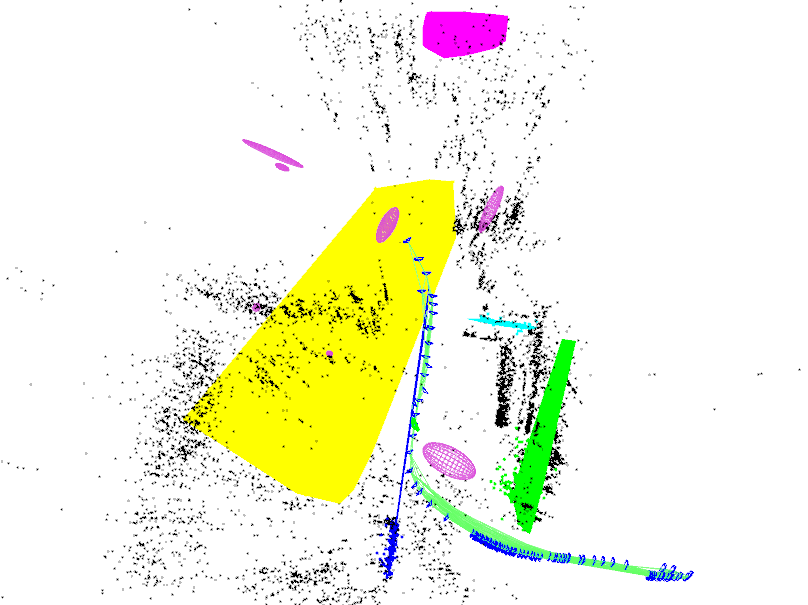}}\\
\caption{Qualitative results for different TUM and NYUv2 datasets. The sequences vary from rich planar structures to multi-object cluttered office scenes}
\label{fig:experiments} 
\vspace{-4mm}
\end{figure}

The proposed system is built in C++ on top of 
the state-of-the-art ORB-SLAM2~\cite{orbslam} and utilizes its front-end for tracking ORB features, while the
back-end for the proposed system is implemented in C++ using g2o~\cite{kummerle2011g}.
Evaluation is performed on a commodity machine with Intel Core i7-4790 processor and a single GTX980 GPU card in near 20 fps and carried out on publicly available TUM \cite{tum-dataset}, NYUv2 \cite{nyuv2-dataset}, and KITTI \cite{kitti} datasets that contain rich planar low-texture scenes to multi-object offices and outdoor scenes. 
Qualitative and quantitative evaluations are carried out using different mixture of landmarks and comparisons are presented against point-based monocular ORB-SLAM2~\cite{orbslam}. 

\subsection{TUM and NYUv2}\label{subsec:tumnyu}
Qualitative evaluation on TUM and NYUv2 for sequences \texttt{fr2/desk}, \texttt{nyu/office\_1b}, and \texttt{nyu/nyu\_office\_1} is illustrated in Fig.~\ref{fig:experiments} for different scenes and landmarks. 
Columns~(a)-(d) show 
the image frame with tracked features and possible detected objects, 
detected and segmented planes, and the reconstructed map from two different viewpoints, respectively. 
For some low or no texture sequences in TUM and NYUv2 datasets point-based SLAM system fail to track the camera, 
however the present rich planar structure is exploited by our system along with the Manhattan constraints to yield more accurate trajectories and semantically meaningful maps. 

The reconstructed maps are semantically rich and consistent with the ground truth 3D scene, for instance in \texttt{fr2/desk}, with presence of all landmarks and constraints, the map consists of planar monitor orthogonal to the desk, and quadrics corresponding to objects are tangent to the supporting desk, congruous with the real scene. Red ellipses in Fig.~\ref{fig:experiments} column~(a) are the projection of their corresponding quadric objects in the map. Further evaluations can be found in the supplemental video.

\begin{figure}[t]
\centering
\setcounter{subfigure}{0}
\subfloat[PlaneNet detector]{\includegraphics[width=0.15\textwidth]{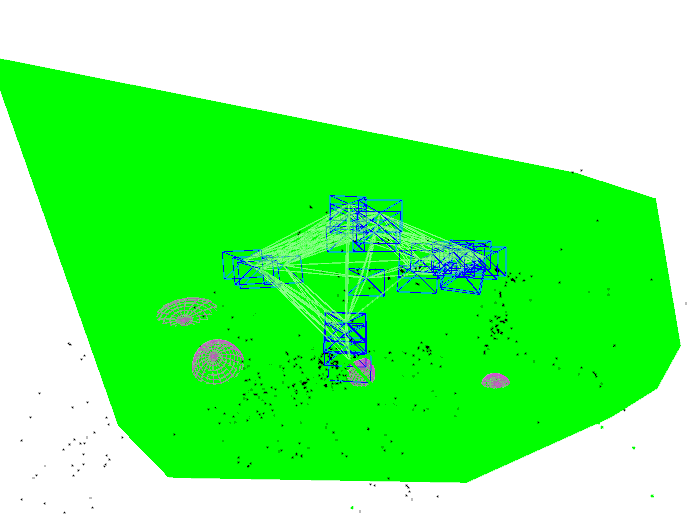}}~
\subfloat[Proposed detector]{\includegraphics[width=0.15\textwidth]{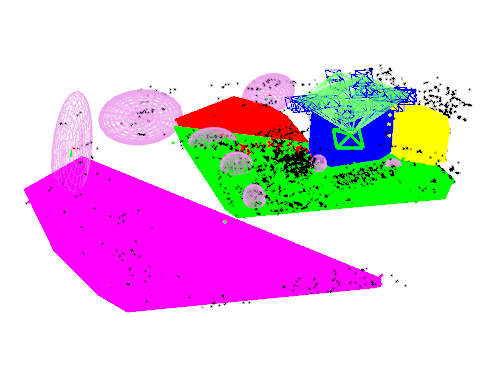}}~
\subfloat[Baseline detector]{\includegraphics[width=0.15\textwidth]{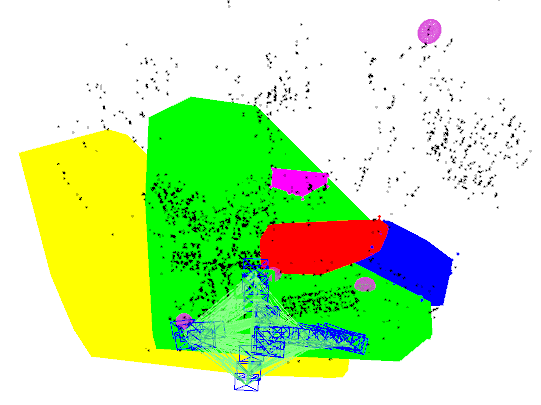}}
\caption{Qualitative comparison of using different plane detectors in our monocular SLAM system for \texttt{fr1/xyz}.}
\label{fig:plane_detector_experiment} 
\end{figure}

One of the main reasons for the improved accuracy of camera trajectory and consistency of the global map is the addressing of subtle but extremely important problem of scale drift. 
In a monocular setting, the estimated scale of the map can change gradually over time. 
In our system, the consistent metric scale of the planes (from CNN) and the presence of point-plane constraints allow observation of the absolute scale, which can further be improved by adding priors about the extent of the objects represented as quadrics.

\begin{figure}[bt]
\centering
\subfloat{\includegraphics[width=\columnwidth]{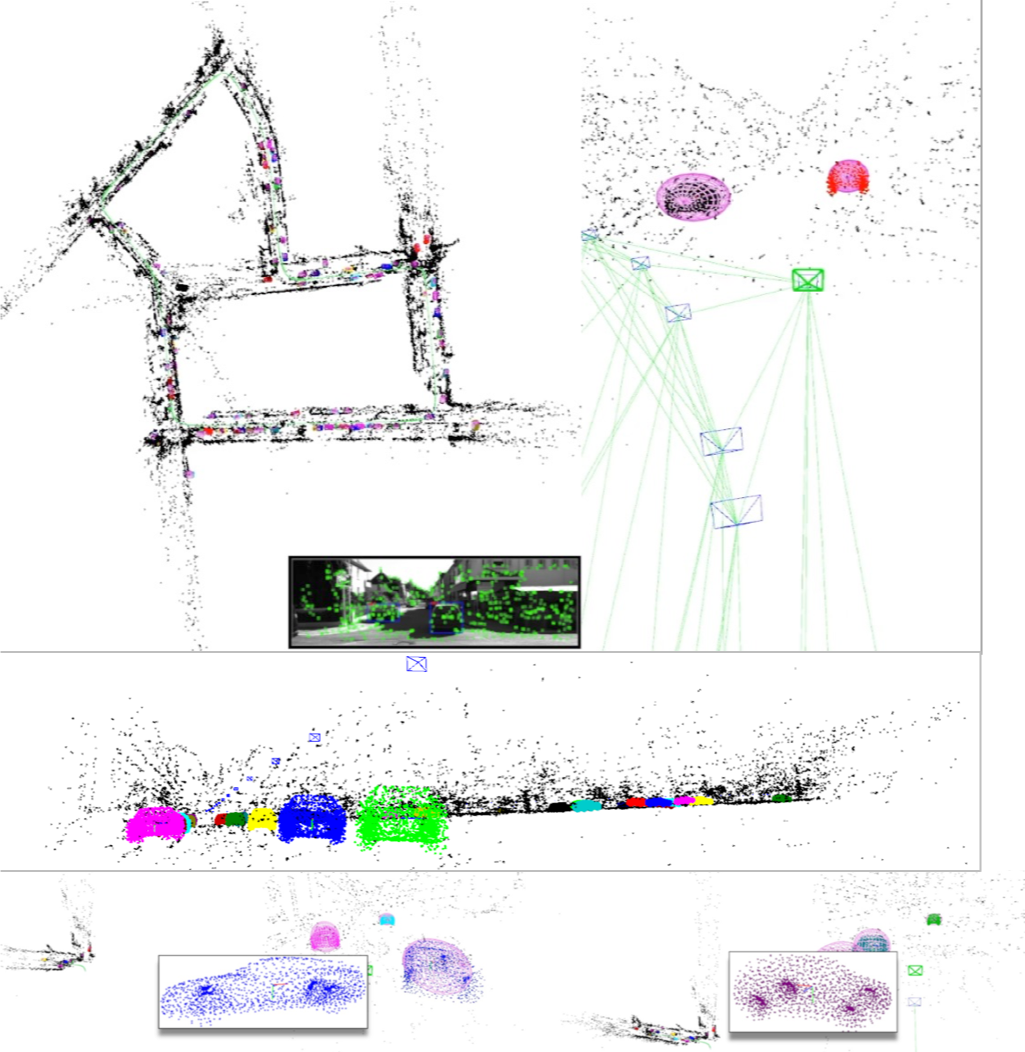}}
\caption{Reconstructed map and camera trajectories for \texttt{KITTI-7} with our SLAM. Proposed object observation and shape prior factors are effective in this reconstruction. 
The reconstructed point-cloud models for a \texttt{sedan} and \texttt{hatchback} car parked beside the road are rendered along with the quadrics
}
\label{fig:kitti_experiment} 
\end{figure}

\begin{table}[b]
	\centering
    \caption{RMSE (\texttt{cm}) of ATE for our monocular SLAM against monocular ORB-SLAM2. Percentage of improvement over ORB-SLAM2 is represented in [~]. See \ref{subsec:tumnyu}}
    \resizebox{\columnwidth}{!}{
    \begin{tabular}{l |c|c|c|c|c|c}
	\hline
          Dataset            		&    \# KF		& ORB-SLAM2&         PP   			&    	        PP+M					  &	      PO 		  	  &		  PPO+MS      
          \\\hline
	 \texttt{fr1/floor}      		&	125			& 1.7971   & 		1.6923          &  \textbf{1.6704} \scriptsize{[7.05\%]}  & 		 ---          &        ---   	   \\\hline 
	 \texttt{fr1/xyz}        		&	30			& 1.0929   & 		1.0291          & 		0.9802  	& 	   	1.0081        &  \textbf{0.9680} \scriptsize{[11.43\%]}  \\\hline 
	 \texttt{fr1/desk}      		&	71			& 1.3940   & 		1.2961  		&  		1.2181  	& 		1.2612   	  &  \textbf{1.2126} \scriptsize{[13.01\%]}  \\\hline 
	 \texttt{fr2/xyz}      			&	28			& 0.2414   & 		0.2213          & 		0.2189    	& 		0.2243   	  &  \textbf{0.2179} \scriptsize{[9.72\%]}  \\\hline 
	 \texttt{fr2/rpy}      			&	12			& 0.3728   & 		0.3356          & 		0.3354    	& 		0.3473        &  \textbf{0.3288} \scriptsize{[11.79\%]}  \\\hline 
	 \texttt{fr2/desk}      		&	111			& 0.8019   & 		0.7317          & 		0.7021    	& 		0.7098        &  \textbf{0.6677} \scriptsize{[16.74\%]}  \\\hline 
	 \texttt{fr3/long\_office}    	&	193			& 1.0697   & 		0.9605 			&  		0.9276  	& 	   	0.9234   	  &  \textbf{0.8721} \scriptsize{[18.47\%]} \\\hline 
	\end{tabular}
    }
	\label{tab:errors}
\end{table}

One of the important factors that can affect the system performance is the quality of estimated plane parameters.
Reconstructed maps are shown in Fig.~\ref{fig:plane_detector_experiment} for two different monocular plane detectors incorporated in our system: 
\textbf{a)} PlaneNet~\cite{plane-net}, 
\textbf{b)} our proposed plane detector (See Section \ref{sec:mono_plane}).
Baseline comparison is made against a depth based plane detector that uses connected component segmentation of the point cloud (\cite{trevor2013efficient, mehdi-arxiv}). 
The detected planes are then used in the monocular system for refinement.
As seen in Fig.~\ref{fig:plane_detector_experiment}(a) PlaneNet only captures the planar table region successfully and fails for the other regions.
The proposed detector captures the monitors on the table shown in column (b), however it misses the monitor behind and also reconstructs the two same height tables with a slight vertical distance. 
As shown in Fig.~\ref{fig:plane_detector_experiment}(c) the baseline plane detector captures the smaller planar regions more accurately and same height tables as one plane, as expected because of using additional \textit{depth} information.
Table~\ref{tab:errors_plane_detectors} reports the comparison of these three approaches for plane detection in different sequences of TUM datasets. 
It can be seen that the depth based detector is the most informative, however the proposed method is better than PlaneNet in most cases.

We perform an ablation study 
to demonstrate the efficacy of introducing various combinations of the proposed landmarks and constraints.
The RMSE of Absolute Trajectory Error (ATE) is reported in Table \ref{tab:errors}. Estimated trajectories and ground-truth are aligned using a similarity transformation \cite{horn}.
In the first case, points are augmented with planes (\texttt{PP}) and constraint for points and corresponding planes is included. This already improves the accuracy over baseline and imposing additional Manhattan constraint in the second case (\texttt{PP+M}) improves ATE even further. In these two cases the error is significantly reduced by first exploiting the structure of the scene and second by reducing the scale-drift, as discussed earlier, using metric information about planes. 

For the sequences containing common COCO~\cite{coco} objects, the presence of objects represented by quadric landmarks along with points is explored in the third case (\texttt{PO}). This case demonstrates the effectiveness of integrating objects in the SLAM map.
Finally, the performance of our full monocular system (\texttt{PPO+MS}) is detailed in the last right column of Table \ref{tab:errors} with the presence of all landmarks points, planes, and objects and also Manhattan and supporting/tangency constraints. This case shows an improvement against the baseline in all of the evaluated sequences, in particular for \texttt{fr3/long\_office} we have seen a significant decline in ATE (18.47\%) as a result of the presence of a large loop in this sequence, where our proposed multiple-edges for observations of planes and quadric objects in key-frames have shown their effectiveness in the global loop closure.

\subsection{KITTI benchmark}\label{subsec:kitti}
To demonstrate the efficacy of our proposed object detection factor, object tracking, and also shape prior factor induced from incorporated point-cloud (reconstructed by CNN from single-view) in our SLAM system, we evaluate our system on KITTI benchmark. 
For reliable frame-to-frame tracking, we use the stereo variant of ORB-SLAM2, however object detection and plane estimation are still carried out in a monocular fashion.
The reconstructed map with quadric objects and incorporated point-clouds (See Section \ref{subsec:obj_pointcloud}) is illustrated for \texttt{\textbf{KITTI-7}} in Fig.~\ref{fig:kitti_experiment}. The instances of different cars are rendered in different colors. 

\begin{table}[t]
	\centering
    \caption{RMSE for ATE (\texttt{cm}) using different plane detection methods in our monocular SLAM. See \ref{subsec:tumnyu}}
	\begin{tabular}{l c|c|c}
	\hline
          Dataset            		&    PlaneNet~\cite{plane-net} 		&         Proposed Detector             &       Baseline		\\\hline
	  \texttt{fr1/xyz}        		& 			0.9701   				& 		   \textbf{0.9680}             	&		0.8601			\\\hline 
	  \texttt{fr1/desk}      		& 			1.2191  	 			& 		   \textbf{1.2126}              &		1.0397			\\\hline 
	  \texttt{fr2/xyz}      		& 			0.2186   				& 		    \textbf{0.2179}             &		0.2061			\\\hline 
	  \texttt{fr1/floor}      		& 		\textbf{1.6562} 			& 			    1.6704                  &		1.4074			\\\hline 
	\end{tabular}
	\label{tab:errors_plane_detectors}
\end{table}

\section{CONCLUSIONS}\label{sec:conclusions}
This work introduced a monocular SLAM system that can incorporate learned priors in terms of plane and object models in an online real-time capable system. We show that introducing these quantities in a SLAM framework allows for more accurate camera tracking and a richer map representation without huge computational cost. This work also makes a case for using deep-learning to improve the performance of traditional SLAM techniques by introducing higher level learned structural entities and priors in terms of planes and objects.

\section*{ACKNOWLEDGMENT}
This work was supported by ARC Laureate Fellowship FL130100102 to IR and the ARC Centre of Excellence for Robotic Vision CE140100016.
%

\bibliographystyle{IEEEtran}
\bibliography{IEEEabrv,refs}


\end{document}